\documentclass{article}
\usepackage{spconf,amsmath,graphicx,hyperref}
\usepackage{multirow}
\usepackage{algorithm}
\usepackage{algorithmic}
\usepackage{booktabs}
\usepackage{float}
\usepackage{stfloats}

\title{An Effective Data Augmentation Method by Asking Questions about Scene Text Images}
\name{Xu Yao, Lei Kang}
\address{Computer Vision Center, Barcelona, Spain\\
\texttt{\{xyaochen7@alumnes.ub.edu, lkang@cvc.uab.es\}}}
\begin{document}
%
\maketitle
\begin{abstract}
Scene text recognition (STR) and handwritten text recognition (HTR) face significant challenges in accurately transcribing textual content from images into machine-readable formats. Conventional OCR models often predict transcriptions directly, which limits detailed reasoning about text structure.
We propose a VQA-inspired data augmentation framework that strengthens OCR training through structured question-answering tasks. For each image–text pair, we generate natural-language questions probing character-level attributes such as presence, position, and frequency, with answers derived from ground-truth text. These auxiliary tasks encourage finer-grained reasoning, and the OCR model aligns visual features with textual queries to jointly reason over images and questions.
Experiments on WordArt and Esposalles datasets show consistent improvements over baseline models, with significant reductions in both CER and WER. Our code is publicly available at \url{https://github.com/xuyaooo/DataAugOCR}.

\end{abstract}
\section{Introduction}
\label{sec:intro}

Scene Text Recognition (STR) and Handwritten Text Recognition (HTR) are pivotal tasks in optical character recognition (OCR), aiming to convert textual content in images into machine-readable formats.

We propose a complementary strategy by framing OCR as a visual question answering (VQA) problem. Instead of treating OCR solely as predicting entire words, we generate auxiliary questions probing character-level properties. For example, given ground-truth transcription, the model answers questions such as \textit{``What is the second character?''} or \textit{``How many times does `L' appear?''}. Each question provides fine-grained supervision beyond word-level recognition.

Unlike traditional augmentation that modifies images, our approach enriches supervision by generating multiple character-level questions for each image-text pair, creating diverse reasoning pathways without additional visual data.

Our contributions are: \textbf{i)} VQA-based OCR augmentation introducing a novel paradigm converting training samples into multiple question-answering tasks. \textbf{ii)} Structured question taxonomy providing systematic character-level questions with probabilistic sampling for diverse supervision. \textbf{iii)} Empirical validation showing consistent improvements on WordArt and Esposalles datasets without requiring additional data.

\section{Related Work}
\label{sec:format}

STR focuses on recognizing text in natural scenes, such as street signs and advertisements captured in uncontrolled environments. The field faces significant challenges due to reliance on synthetic training datasets like MJSynth and SynthText, while evaluation occurs on smaller real-world benchmarks. This creates a substantial domain gap where models struggle with real-world variations. Recent efforts have focused on developing STR-specific data augmentation techniques to bridge this synthetic-to-real distribution shift \cite{Atienza_2021_ICCV}.

HTR addresses the recognition of handwritten text, which introduces additional complexities due to substantial variability in handwriting styles across different writers. Deep learning models require extensive diverse training data to achieve robust generalization across unseen handwriting styles. Current HTR datasets such as IAM are considerably smaller than standard computer vision benchmarks, leading to overfitting. To mitigate this data limitation, researchers have developed augmentation strategies that generate synthetic training samples \cite{8270041, 7780622}.

Visual Question Answering (VQA) has emerged as a fundamental task in multi-modal learning, requiring models to understand both visual content and natural language to provide accurate answers. Traditional VQA systems primarily focus on natural images and general visual reasoning tasks\cite{7410636,Malinowski2014AMA}. However, a specialized branch of VQA has developed to address text-rich scenarios where textual information in images plays a crucial role in answering questions \cite{8978122,singh2019vqamodelsread}.

Inspired by the Instruction-Guided Scene Text Recognition (IGTR) framework \cite{10820836}, we explore applying question-answering paradigms to enhance text recognition. IGTR formulates STR as an instruction learning problem and understands text images by predicting character attributes through $\langle$condition, question, answer$\rangle$ instruction triplets.

Our approach builds upon IGTR but introduces several key innovations. While IGTR employs conditional instruction triplets, we propose a systematic five-category question taxonomy that focuses purely on character-level attribute queries without conditional statements. We further introduce a probabilistic sampling strategy that selects question categories according to specified probabilities, allowing flexible emphasis across categories.

\section{Proposed Method}
\label{sec:pagestyle}

\subsection{Problem Formulation}
\label{ssec:Problem Formulation}

Let $\{X,Y\}$ be a text dataset with images $X=\{I_n\}_{n=1}^N$ and corresponding labels $Y=\{y_n\}_{n=1}^N$, where each $y_n$ is a character sequence for image $I_n$. The standard OCR task learns a model $f: X \rightarrow Y$ to map $I \rightarrow y$, where $X$ is the image space and $Y$ is the text sequence space.

Our framework extends this formulation by introducing three interconnected spaces: Image Space $X$ (the space of scene text images), Question Space $\mathcal{Q}$ (the space of questions about text content), and Answer Space $\mathcal{A}$ (the space of corresponding answers to questions in $\mathcal{Q}$).

We define a question-answer generation function $g$ that maps each image-text pair $(I, y)$ to a set of question-answer pairs: $g(I, y) = \{(q_i, a_i)\}_{i=1}^K$, where each $(q_i, a_i) \in \mathcal{Q} \times \mathcal{A}$ probes different aspects of the text sequence $y$ (e.g., character presence, position...).

Additionally, we learn a question-answering model $h: (\mathcal{Q} \times X) \rightarrow \mathcal{A}$ that maps question-image pairs to answers:

$$h(q, I) = a$$

For inference, note that the standard OCR task can be viewed as a special case where the question is ``What is this word?'', making the usual recognition model $f: X \rightarrow Y$ equivalent to $h(\text{``What is this word?''}, I)$.
\subsection{Architecture}
\label{ssec:Architecture}

\begin{figure*}[t]  
\centering
\includegraphics[width=0.85\textwidth]{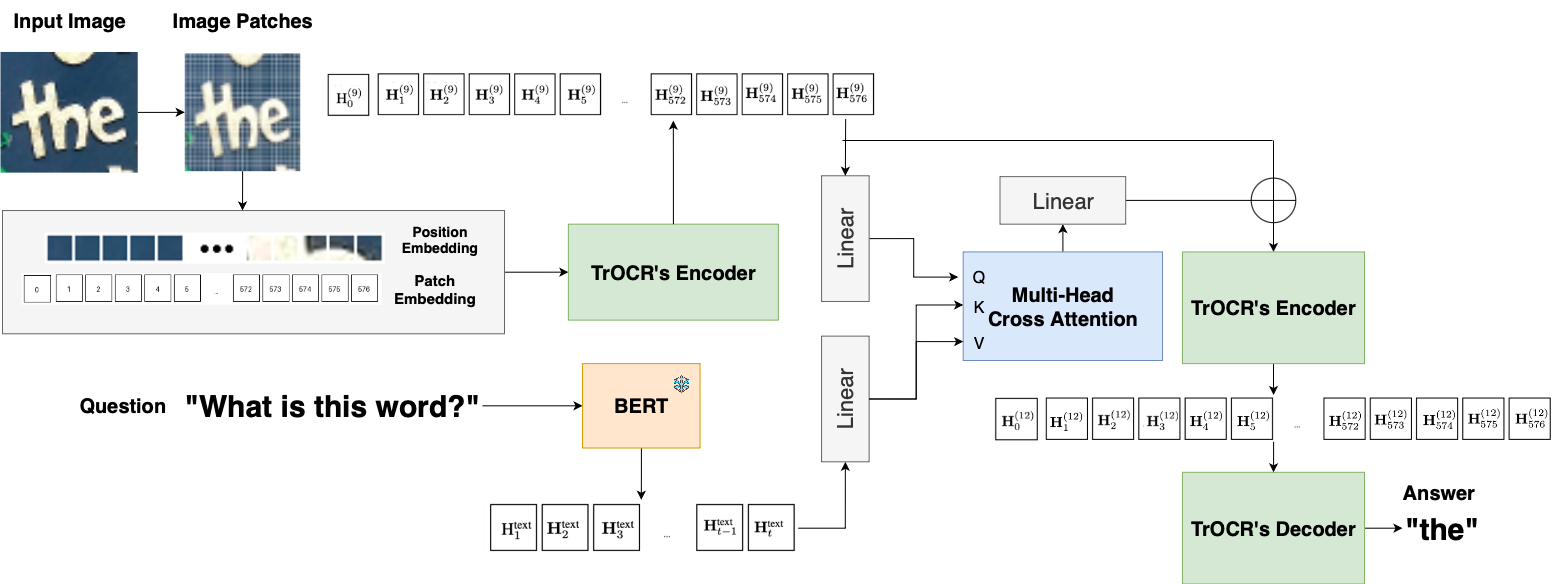}  
\caption{Overall architecture of the proposed VQA-based OCR framework showing the integration of visual features from TrOCR encoder with textual query embeddings through cross-modal attention. }
\label{fig:arquitecture}
\end{figure*}

The proposed method’s architecture is shown in Figure \ref{fig:arquitecture}.

Our architecture builds upon the TrOCR foundation \cite{TrOCR} by introducing a cross-modal attention mechanism that allows visual feature extraction to be conditioned on textual queries. Following the original TrOCR-Base architecture, we employ a Vision Transformer (BEiT \cite{BEiT}) with 12 encoder layers as our visual backbone. Input images are resized to $384 \times 384$ pixels and divided into $16 \times 16$ patches with positional embeddings, creating visual tokens $\mathbf{H}^{(0)}$ of shape $577 \times 768$ (576 patches plus one CLS token). For textual input processing, we utilize a frozen pre-trained BERT-base-uncased model \cite{BERT} to generate contextualized embeddings. Given a textual query $T$, BERT produces textual features $\mathbf{H}^{\text{text}}$.

The key innovation is the strategic insertion of a cross-modal attention module after the 9th transformer block. Both visual and textual features are fed into fully connected layers to reduce their dimensions from 768 to $d_{\text{cross}}$ for both visual features from $\mathbf{H}^{(9)}$ and textual features from $\mathbf{H}^{\text{text}}$. The multi-head attention mechanism from \cite{attentionIsAllYouNeed} uses four attention heads where the query is derived from the reduced visual features, while both key and value are derived from the reduced textual features. The cross-attention output is integrated back into the visual processing stream through a residual connection and projected back to dimension 768, and the enhanced visual features are processed through the remaining transformer blocks to produce the final text-guided visual representation $\mathbf{H}^{(12)}$.

The enhanced visual features $\mathbf{H}^{(12)}$ are fed into the standard TrOCR decoder, which uses 12 layers of the RoBERTa decoder \cite{liu2019robertarobustlyoptimizedbert}, generating character sequences auto-regressively.

\subsection{Question Taxonomy and Generation Strategy}
\label{ssec:question_taxonomy}

Our VQA-based augmentation relies on a systematic taxonomy of character attribute questions organized into five categories, each containing two specific subcategories. This structure ensures comprehensive coverage of character-level reasoning capabilities. The taxonomy is designed to decompose the complex OCR task into interpretable sub-problems that can be learned more effectively through multi-task learning. The question-answer generation function $g$ introduced in Section \ref{ssec:Problem Formulation} generates questions from this taxonomy.

The five categories include: (1) Recognition questions for standard OCR output, (2) Character Presence Analysis questions, (3) Positional Analysis questions, (4) Structural Analysis questions, and (5) Boundary Analysis questions. Table \ref{tab:question_examples} illustrates the complete taxonomy with concrete examples for the word "HELLO" and shows the different types of question-answer pairs generated by the function $g$:

\begin{table*}[h]
\centering
\caption{Question taxonomy with examples for the word "HELLO"}
\label{tab:question_examples}
\begin{tabular}{|l|l|l|l|l|}
\hline
\textbf{Category} & \textbf{Subcategory} & \textbf{Question Example} & \textbf{Answer} & \textbf{Answer Type} \\
\hline
\multirow{1}{*}{Recognition (0)} & Base OCR & What is this word? & HELLO & Text \\
\hline
\multirow{2}{*}{Presence (1)} & Existence & Is the character 'L' in this word? & Yes & Binary \\
& Frequency & How many times does 'L' appear? & 2 & Numerical \\
\hline
\multirow{2}{*}{Positional (2)} & Position & What is the character at position 2? & E & Character \\
& Relation & Does 'E' come before 'H' in this word? & No & Binary \\
\hline
\multirow{2}{*}{Structural (3)} & Length & What is the total number of characters? & 5 & Numerical \\
& Repetition & Is there any repeated character? & Yes & Binary \\
\hline
\multirow{2}{*}{Boundary (4)} & Start & Does this word start with 'H'? & Yes & Binary \\
& End & Does this word end with 'O'? & Yes & Binary \\
\hline
\end{tabular}
\end{table*}

\subsection{Probabilistic Question Sampling}
\label{ssec:probabilistic_sampling}
Our training methodology employs a probabilistic sampling strategy to determine which question categories to include for each training sample. This approach ensures diverse question exposure while maintaining computational efficiency by controlling the number of questions generated per image-text pair.

To determine optimal sampling probabilities, we conduct systematic ablation experiments evaluating each category's contribution. Our experimental design tests combinations where recognition questions (Category 0) are paired with each major attribute category: Recognition + Presence, Recognition + Positional, Recognition + Structural, and Recognition + Boundary.

Based on these results, our training uses a controlled sampling mechanism where each sample contains the base recognition question to maintain the primary OCR objective, plus both questions from exactly one attribute category selected with probabilities $\{p_1, p_2, p_3, p_4\}$ where $\sum_{i=1}^{4} p_i = 1$. The probability distribution reflects empirical performance from the ablation study: the two most effective categories receive 30\% probability each, the third-most effective receives 25\%, and the least effective receives 15\%, with each selected category contributing its two question types as shown in Table~\ref{tab:question_examples}.

\section{Experiments}
\label{sec:Experiments}

\subsection{Datasets and Performance Metrics}
\label{ssec:Datasets and Performance Metrics}

We evaluate our method on two diverse datasets to demonstrate its effectiveness across different text recognition scenarios:

WordArt: This dataset from \cite{xie2022understandingwordartcornerguidedtransformer} focuses on artistic scene text recognition and contains 4,805 training images and 1,511 validation images. The images feature diverse artistic text styles including posters, greeting cards, covers, and billboards, making recognition challenging due to the variety of fonts, colors, and artistic effects.

Esposalles: From \cite{esposalles}, we use the word-level recognition subset containing 21,786 training images and 8,026 test images. The training data is organized in three separate parts, totalling those 21,786 images This dataset consists of historical handwritten marriage records, presenting challenges such as varied writing styles, ink fading, and document degradation.

For evaluation, we employ the standard Character Error Rate (CER) and Word Error Rate (WER) metrics. Both metrics are reported as percentages (0-100), where lower values indicate better performance.

\subsection{Implementation Details}
\label{ssec:Implementation Details}

All experiments used PyTorch \cite{pytorch} on an NVIDIA TITAN Xp GPU (12GB). Images follow standard TrOCR preprocessing: 384×384 pixels with 16×16 patches. We train our model using the AdamW \cite{loshchilov2018decoupled} optimizer with a learning rate of 5e-6, effective batch size of 6, and weight decay of 1e-2, and StepLR scheduling (step size 5, gamma 0.9). We apply gradient clipping with a maximum norm of 2.0 to stabilize training. A dropout rate of 0.1 is applied to all dropout layers.

For our cross-modal attention module, we set the reduced dimension $d_{\text{cross}}$ to 384 with 4 attention heads, inserted after the 9th transformer block. We train for up to 50 epochs and select the model checkpoint with the best validation performance based on Word Error Rate (WER) on the test set.

For initialization, we use different TrOCR pretrained models depending on the dataset: trocr-base-str (fine-tuned on IC13, IC15, IIIT5K, SVT) for the WordArt dataset, and trocr-base-handwritten (fine-tuned on the IAM dataset) for the Esposalles dataset.

\subsection{Ablation Studies}
\label{ssec:Ablation Studies}

Table \ref{tab:ablation_combined} shows the performance of different question category combinations for both WordArt and Esposalles datasets. All ablation experiments are conducted for 10 epochs. For computational efficiency, ablation experiments on Esposalles only use Part 2 of the training data (9,624 samples).

\begin{table*}[!t]
\centering
\caption{Ablation study results on WordArt and Esposalles datasets}
\label{tab:ablation_combined}
\begin{tabular}{lccccc}
\toprule
& \multicolumn{2}{c}{\textbf{WordArt}} & & \multicolumn{2}{c}{\textbf{Esposalles}} \\
\cmidrule{2-3} \cmidrule{5-6}
\textbf{Method} & \textbf{WER (\%)} & \textbf{CER (\%)} & & \textbf{WER (\%)} & \textbf{CER (\%)} \\
\midrule
Recognition + Presence & 27.92 & 11.36 & & 4.84 & 1.51 \\
Recognition + Positional & 27.72 & 11.79 & & 5.03 & 1.47 \\
Recognition + Structural & 28.08 & 12.73 & & 4.82 & 1.42 \\
Recognition + Boundary & 29.05 & 15.62 & & 5.40 & 1.68 \\
\bottomrule
\end{tabular}
\end{table*}

The results show that different question categories provide varying improvements over the baseline. Based on these empirical findings, we determine the optimal sampling probabilities for each dataset as described in Section \ref{ssec:probabilistic_sampling}. For WordArt, we assign probabilities $\{p_1=0.3, p_2=0.3, p_3=0.25, p_4=0.15\}$ to categories \{Presence, Positional, Structural, Boundary\} respectively. For Esposalles, the probabilities are $\{p_1=0.3, p_2=0.25, p_3=0.3, p_4=0.15\}$, reflecting the different category effectiveness patterns observed in historical handwritten text recognition.

\subsection{STRaug Baseline Implementation}
\label{ssec:STRaug Baseline}

To provide a comprehensive evaluation against established augmentation techniques, we implement STRAug \cite{Atienza_2021_ICCV} as an additional baseline. STRAug introduces 36 specialized image augmentation functions, organized into 8 logical groups: Warp, Geometry, Noise, Blur, Weather, Camera, Pattern, and Process.
STRAug follows the RandAugment framework \cite{9150790}, which uses two key hyperparameters: N (the number of transformations groups applied sequentially to each image, selected with uniform
probability) and M (the magnitude controlling the intensity of each transformation).

For our implementation, we configure STRaug with N=2, magnitude M uniformly sampled from \{0,1,2\}, and 50\% transformation application probability per training sample. This configuration follows the original paper's recommendations.

\subsection{Results and Comparison}
\label{ssec:Results and Comparison}

We evaluate our VQA-augmented approach against baseline TrOCR models and TrOCR with STRaug augmentation.

\begin{table}[h]
\centering
\caption{Comprehensive results on WordArt dataset}
\label{tab:comprehensive_wordart}
\begin{tabular}{lcc}
\toprule
\textbf{Method} & \textbf{WER (\%)} & \textbf{CER (\%)} \\
\midrule
TrOCR (base-str) & 30.64 & 12.76 \\
TrOCR + STRaug & 29.84 & 12.32 \\
\midrule
Ours (VQA-augmented) & \textbf{27.26} & \textbf{11.38} \\
\bottomrule
\end{tabular}
\end{table}

\begin{table}[h]
\centering
\caption{Comprehensive results on Esposalles dataset}
\label{tab:comprehensive_esposalles}
\begin{tabular}{lcc}
\toprule
\textbf{Method} & \textbf{WER (\%)} & \textbf{CER (\%)} \\
\midrule
TrOCR (base-handwritten) & 11.95 & 5.65 \\
TrOCR + STRaug & 10.91 & 4.95 \\
\midrule
Ours (VQA-augmented) & \textbf{3.80} & \textbf{1.10} \\
\bottomrule
\end{tabular}
\end{table}

As shown in Tables \ref{tab:comprehensive_wordart} and \ref{tab:comprehensive_esposalles}, our VQA-based augmentation achieves consistent improvements across both datasets. Our approach outperforms both baselines with substantial reductions in both CER and WER. These gains validate that enriching supervision through character-level reasoning questions provides an effective alternative to traditional visual augmentation techniques for enhancing OCR performance.

\section{Conclusion}
\label{sec:conclusion}

We presented a VQA-based data augmentation framework that enhances OCR training by converting image-text pairs into structured question-answering tasks. By incorporating semantic meaning through questions, the model is compelled to learn the semantic information present in scene or handwritten text images, making it an effective method for data augmentation that goes beyond traditional visual transformations. Experiments demonstrate consistent improvements over baseline models, showing that enriched supervision through character-level questions offers a promising direction for advancing text recognition systems.

\vfill\pagebreak

\newpage

\small

\section{Acknowledgments}

This work has been supported by the Beatriu de Pin\'{o}s del Departament de Recerca i Universitats de la Generalitat de Catalunya (2022 BP 00256), European Lighthouse on Safe and Secure AI (ELSA) from the European Union's Horizon Europe programme under grant agreement No.~101070617, and Riksbankens Jubileumsfond, grant M24-0028 (Echoes of History: Analysis and Decipherment of Historical Writings, DESCRYPT).

\bibliographystyle{IEEEbib}
\bibliography{refs}

\end{document}